\documentclass[letterpaper, 10 pt, conference]{ieeeconf}  

\IEEEoverridecommandlockouts                              

\usepackage{hyperref}
\usepackage{multirow}
\usepackage{romannum}
\usepackage{graphicx}
\usepackage{amsmath}
\usepackage{amssymb}
\usepackage{breqn}
\usepackage{subcaption}
\usepackage{balance}

\usepackage{pgfplots}
\usepackage{booktabs, dcolumn}

\pgfplotsset{compat=newest}
\pgfplotsset{plot coordinates/math parser=false}
\newlength\figureheight
\newlength\figurewidth

\makeatletter
\let\NAT@parse\undefined
\makeatother
\usepackage[numbers,sort,compress]{natbib}
\usepackage{hyperref}
\usepackage{xcolor,colortbl}

\title{\bf
Self-supervised Object Tracking with Cycle-consistent Siamese Networks
}

\author{Weihao Yuan, Michael Yu Wang, and Qifeng Chen 
\thanks{Authors are with the Hong Kong University of Science and Technology, Hong Kong SAR, China. W. Yuan (\href{mailto:weihao.yuan@connect.ust.hk}{weihao.yuan@connect.ust.hk}) is with the Department of Electronic and Computer Engineering. M. Y. Wang is with the Department of Mechanical and Aerospace Engineering and the Department of Electronic and Computer Engineering. Q. Chen (\href{mailto:cqf@ust.hk}{cqf@ust.hk}) is with the Department of Computer Science and Engineering and the Department of Electronic and Computer Engineering.}
}

\begin{document}

\maketitle


\begin{abstract}


Self-supervised learning for visual object tracking possesses valuable advantages compared to supervised learning, such as the non-necessity of laborious human annotations and online training. In this work, we exploit an end-to-end Siamese network in a cycle-consistent self-supervised framework for object tracking. Self-supervision can be performed by taking advantage of the cycle consistency in the forward and backward tracking. To better leverage the end-to-end learning of deep networks, we propose to integrate a Siamese region proposal and mask regression network in our tracking framework so that a fast and more accurate tracker can be learned without the annotation of each frame. The experiments on the VOT dataset for visual object tracking and on the DAVIS dataset for video object segmentation propagation show that our method outperforms prior approaches on both tasks.

\end{abstract}


\section{INTRODUCTION}


Visual object tracking is an essential task for numerous applications such as autonomous driving \cite{choi2015near}, robotic manipulation \cite{choi2010real}, and video surveillance \cite{tang2017multiple}. Given the position of a target object in the first frame of a video, the task is to estimate its location in subsequent frames. In most cases, the tracking also needs to be run in real time. Although this problem has been studied by many researchers, state-of-the-art methods still suffer from many visual variations such as occlusion, deformation, motion, and illumination change \cite{pont20172017, kristan2018sixth}.


Recent deep network based methods \cite{tao2016siamese, bertinetto2016fully, li2018high, wang2019fast, li2019siamrpn++} for visual object tracking have dominated most benchmarks and demonstrated advantages over traditional tracking methods \cite{bolme2010visual, kiani2013multi, henriques2014high, danelljan2014accurate, choi2016visual, lukezic2017discriminative} in both accuracy and speed. However, most deep-network-based methods require ground-truth object trajectories for training. The annotation of ground truth is laborious and time-consuming, which limits the size of the training data and their applications in unseen scenarios.

To solve this problem, some researchers are exploring self-supervised learning approaches \cite{wang2019unsupervised} for object tracking. A tracker can be learned without the need for annotation on every frame. Additionally, self-supervised methods make online fine-tuning plausible such that they can be applied to unseen scenarios more easily.
Nevertheless, previous self-supervised methods \cite{wang2019unsupervised} are based on a correlation filter, and the performance is limited. On the other hand, Siamese network based trackers \cite{bertinetto2016fully, li2018high, wang2019fast, li2019siamrpn++} have drawn much attention in the community recently. Siamese trackers formulate the visual object tracking task as learning a similarity response map by cross-correlation between the feature embedding of an exemplar patch and a search image. After the cross-correlation, a region proposal network or a mask regression network can contribute to the accurate prediction of the target.


\begin{figure}[]
\centering
\includegraphics[width=1.0\columnwidth, trim={0cm 0cm 0cm 0cm}, clip]{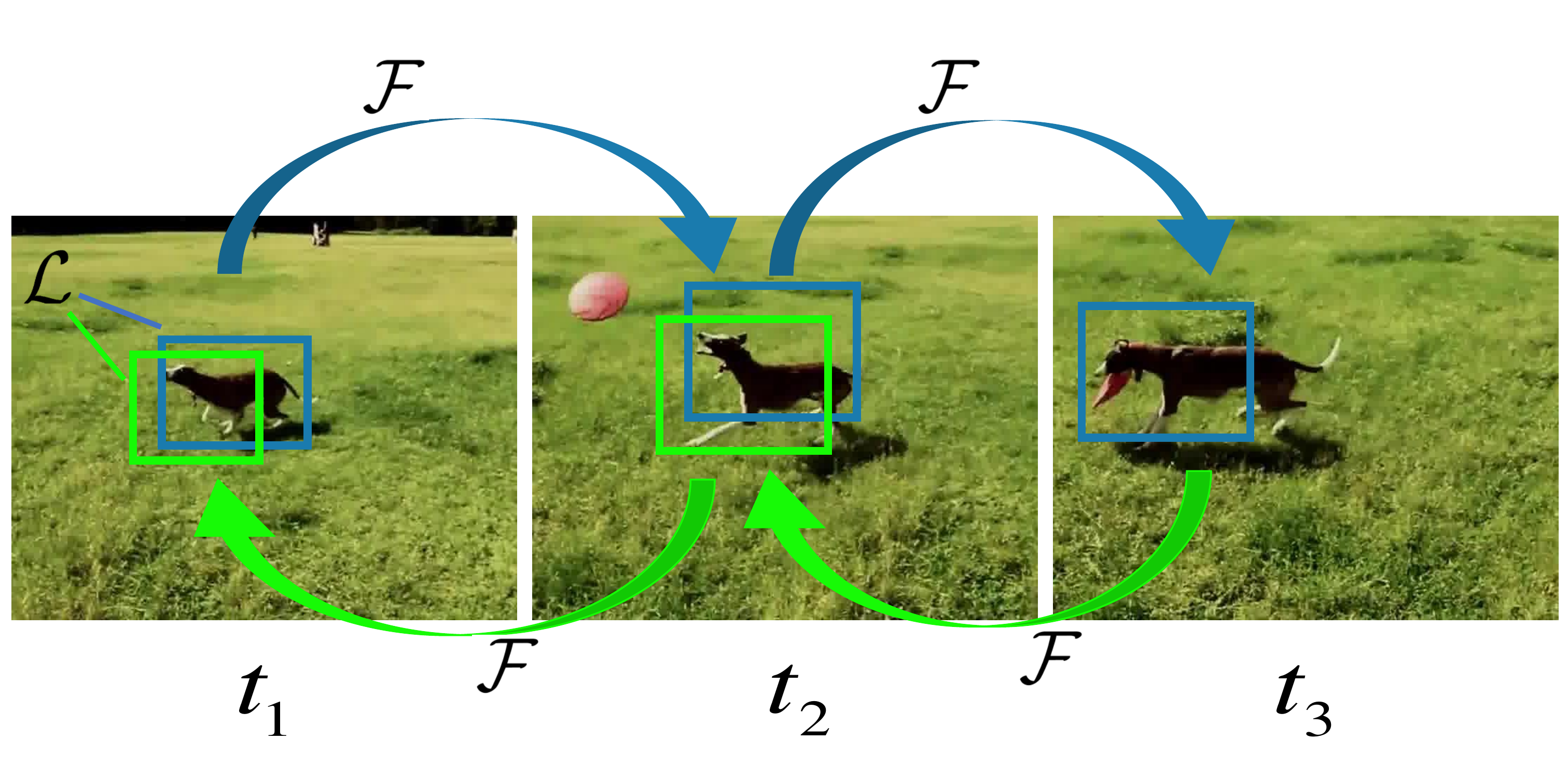}
\caption{Illustration of our self-supervised cycle-consistent framework. Given the bounding box (in blue) of the target object in the first frame, the object is tracked forward (in blue) in subsequent frames and then circularly tracked backward (in green) to the first frame. The discrepancy  between the initial bounding box and the predicted box in the first frame can be used to serve as the supervision to optimize the tracking network $\mathcal{F}(\mathbf{z},\mathbf{x};\theta)$.}
\label{fig:cycle}
\end{figure}

In this paper, we explore self-supervised learning for visual object tracking in a cycle-consistent fashion, utilizing the consistency between forward tracking and backward tracking, as shown in Fig.~\ref{fig:cycle}. For a video sequence, the tracking can be performed from the first frame to the last frame in chronological order, and then in reversed order back to the first frame. If the tracking is accurate, then the estimated location of the target object in the first frame after the backward tracking should be the same as the initial position. Thus, the discrepancy between these two positions can serve as the self-supervision to train the tracking network. 

Following this idea, we introduce an end-to-end Siamese structure with the region proposal network \cite{ren2015faster, li2018high} into the cycle-consistent tracking framework. By leveraging the end-to-end learning ability of deep networks, our framework produces more accurate prediction than previous correlation-filter-based self-supervised methods \cite{wang2019unsupervised}. 
The region proposal network (RPN) after the cross-correlation can estimate more accurate box proposals and improve the performance of the self-supervised tracking framework.

In addition, the cycle-consistent tracking can be also performed at the mask level, which is another fundamental task in computer vision, the video object segmentation propagation (also called semi-supervised video object segmentation) \cite{wen2015jots, perazzi2017learning, cheng2017segflow, bao2018cnn, wang2019learning, lai2019self}. Given the segmentation mask of the objects in the first frame in a video sequence, the task is to predict the segmentation in all the remaining frames.
In our framework, we simply add a mask branch to predict the propagated mask, following the idea in \cite{wang2019fast}. This makes our framework for segmentation propagation still compact and can be run in real time, which is difficult for most approaches for this task, either supervised methods \cite{wen2015jots, perazzi2017learning, cheng2017segflow, bao2018cnn} or self-supervised methods \cite{wang2019learning, lai2019self}. Also, the input of our network is a rough box of the target object rather than the accurate segmentation mask. These advantages mean that our method could be used in more practical applications.

In the experiments, we evaluate our framework on benchmark dataset VOT-2016 and VOT-2018 \cite{kristan2018sixth} for the visual object tracking task, and on DAVIS-2016 and DAVIS-2017 \cite{pont20172017} for the segmentation propagation task. The results show accurate and stable performance of our cycle-consistent framework, with training on unlabeled video sequences. Taking advantage of the tracking target object initialization, our method outperforms previous state-of-the-art self-supervised approaches in both the visual object tracking task \cite{wang2019unsupervised} and the video segmentation propagation task \cite{wang2019learning, lai2019self}, while running in real time.

Our contributions are then summarized as follows.

\begin{enumerate}
\item We introduce the Siamese region proposal network and mask regression module into the cycle-consistent framework to perform better end-to-end self-supervised learning.
\item The proposed method outperforms previous self-supervised algorithms in two tasks of visual object tracking and video segmentation propagation.

\end{enumerate}


\section{RELATED WORK}
\label{sec:related_work}

We first review the methods for visual object tracking in both supervised and self-supervised manners. Then, the approaches for supervised and self-supervised video segmentation propagation are surveyed.

\textbf{Visual Object Tracking. } In the past few years, correlation filter has shown to be fast and effective in comparing the difference between an exemplar image and its searching image due to its transformation in frequency domain, after proposed by Bolme \textit{et al.} \cite{bolme2010visual}. This filter has further been developed by introducing multi-channel \cite{kiani2013multi}, kernel \cite{henriques2014high}, scale estimator \cite{danelljan2014accurate}, attention modular \cite{choi2016visual}, and spatial relationship \cite{lukezic2017discriminative}. Recently deep-feature-based correlation filters \cite{danelljan2017eco, valmadre2017end} have also been proposed for higher accuracy. 

On the other hand, the Siamese structure deep trackers are growing rapidly and have dominated many benchmarks \cite{tao2016siamese, bertinetto2016fully, li2018high, wang2019fast, li2019siamrpn++}. These Siamese network methods formulate visual object tracking as a cross-correlation problem between a template image patch and the searching image, and aim to exploit the mapping ability of deep networks from end-to-end learning. The template patch and the searching image are fed into Siamese networks, and the features are extracted in the same space, after which a cross-correlation is performed to merge two branches to one similarity response map. This structure has been demonstrated to be accurate and fast. To further improve the tracking accuracy, the region proposal network is applied to regress more accurate bounding boxes \cite{li2018high, wang2019fast, li2019siamrpn++}.   

Deep trackers, however, rely heavily on the annotation labels for the training. To address this challenge, some researchers are exploring self-supervised methods for object tracking. 
As a widely-used unsupervised tool, auto-encoder is adopted to extract the generic image feature to detect the moving object in \cite{wang2013learning}. Another important constraint in visual tracking, the forward-backward error, is also widely used to estimate the error for optimizing the trackers \cite{kalal2011tracking}, or labeling the annotations to provide more training data \cite{muller2018trackingnet}.
The forward-backward difference estimating is later extended with the geometry similarity, the cyclic weight, and the appearance similarity \cite{lee2015multihypothesis}.

Recently the forward-backward checking idea has been applied to provide supervision in deep network training \cite{wang2019unsupervised}, which has improved the performance of self-supervised tracking. Nevertheless, there are still few deep trackers trained by self-supervision, and UDT \cite{wang2019unsupervised} only uses a simple discriminative correlation filter to compare the features extracted from the exemplar and search region. Differently, to enhance the self-supervised cycle tracking framework, we use the depth-wise cross-correlation to generate a dense tensor and then feed it into the advanced region proposal network to regress the box location of the target object. In this way, end-to-end self-supervised learning is exploited to improve the cycle tracking framework.

\begin{figure*}[]
\centering
\includegraphics[width=1.8\columnwidth, trim={0cm 0cm 0cm 0cm}, clip]{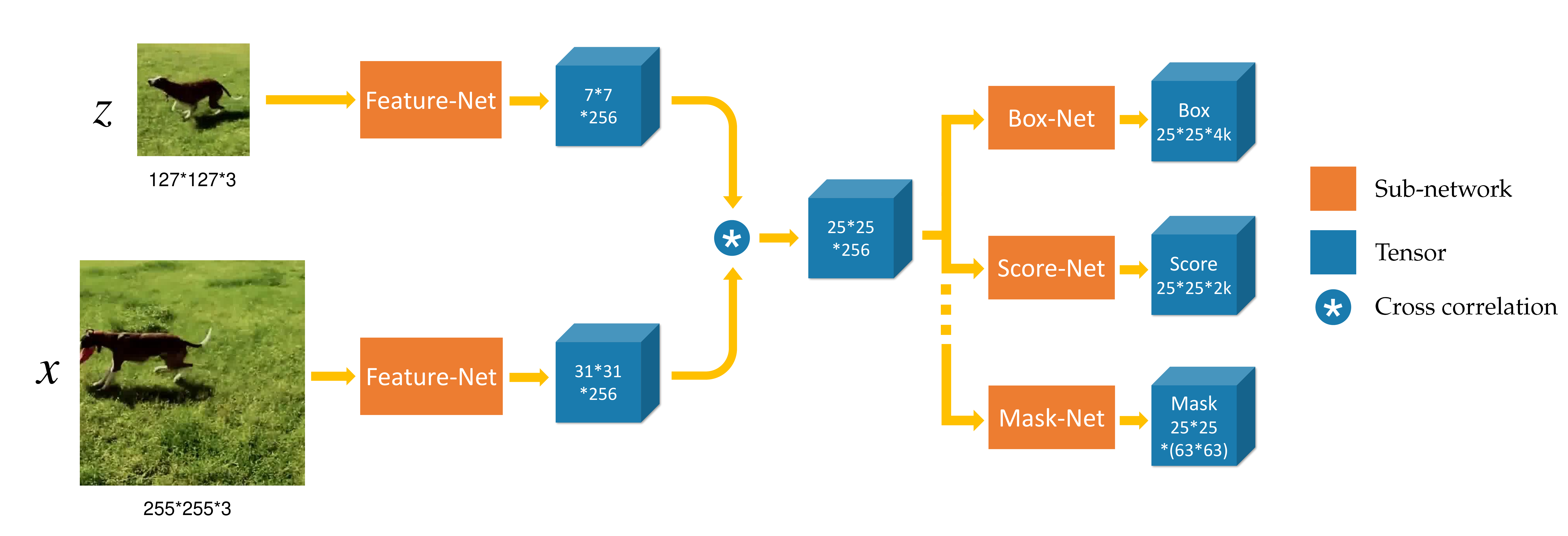}
\caption{The structure of our Siamese tracking network. The input includes two color images, which are the crop of the target object and the image where we need to search for the object. The output is a $25\times25$ dense response map, of which each element includes $k$ box proposals and their corresponding scores. For mask tracking, each element also consists of a flattened mask with a size of $3969=63\times63$.}
\label{fig:network}
\end{figure*}

\textbf{Video Object Segmentation Propagation. } Unlike visual object tracking, video segmentation propagation pays more attention to generating an accurate pixel-level mask for the target objects, such that the algorithms are usually time-consuming and are not in real time.

Traditional methods formulate segmentation propagation as a pixel matching problem. Optical flow is a widely used method to find pixel correspondence between two images but is not accurate for the object mask, so more mid-level features are introduced \cite{liu2010sift, tsai2016object}. Tsai \textit{et al.} improve the optical flow by considering the object and spatial information \cite{tsai2016object}, while SIFT flow considers the correspondence of SIFT features \cite{liu2010sift}.
To find accurate pixel correspondence, graph labeling methods are also widely adopted, where a matching energy function is minimized for correspondence matching \cite{perazzi2015fully, wen2015jots, marki2016bilateral}. To better represent the pixels, deep features are extracted to compare the similarity between pixels \cite{bao2018cnn}. 

Recent methods have tried to directly match the semantic correspondence according to deep features \cite{ufer2017deep, rocco2018end}. Other methods try to process video frames independently \cite{caelles2017one, voigtlaender2017online}. Not relying on temporal information, they fine-tune the trained network using the ground-truth mask provided in the first frame, making it totally a segmentation problem.
Additionally, some methods try to propagate the initial mask from the first frame to subsequent frames once per frame, in a similar way to tracking \cite{wang2019fast}.  

Although the performance of deep learning methods is generally competitive, these approaches often require a large amount of data to do the training. Therefore, some self-supervised works have been proposed \cite{wang2019learning, lai2019self, li2019joint} based on cycle consistency. However, these methods are all first learning visual representations and then performing a correspondence matching. To better leverage the end-to-end learning of deep networks, we propose to add a mask branch in the cycle tracking framework, to learn a segmentation propagation function end-to-end with cycle consistency.

\section{SELF-SUPERVISED TRACKING}

In this section, we first introduce our cycle tracking framework, where the self-supervision works to optimize the tracking system. Then, the network structure and optimization function are given to illustrate how a single forward tracking in this framework is performed.

\subsection{Cycle Object Tracking}

Given a frame $I_{1}$ at time $t_1$, and the patch of the target object $O$ to track, we first forward track the target $O$ to frame $I_{2}$ at another time $t_2$. In the forward tracking, we can get the predicted location and size of the target object $O$ in frame $I_{2}$:

\begin{equation}
    \widetilde{p_2} = \mathcal{F}(p_1, I_2; \theta),
\end{equation}
where $\mathcal{F}$ is the tracking network forwarding with parameters $\theta$, $p_1$ is the patch of the target in frame $I_1$, and $\widetilde{p_2}$ is the predicted patch of the target in frame $I_2$.

As illustrated in Fig.~\ref{fig:cycle}, after the forward tracking, we backward track the patch $\widetilde{p_2}$ into the first frame $I_1$:

\begin{equation}
    \widetilde{p_1} = \mathcal{F}(\widetilde{p_2}, I_1; \theta).
\end{equation}
Then, we can get the predicted patch $\widetilde{p_1}$. Between $\widetilde{p_1}$ and $p_1$ we can calculate a consistency loss. 

In this way, with the original patch as the label, and the predicted patch after cycle tracking as the output, we can obtain the loss to optimize the network parameters without the need for ground-truth annotations.

Furthermore, we can extend the tracking circle to more frames:
\begin{equation}
\begin{aligned}
    \widetilde{p_3} &= \mathcal{F}( \mathcal{F}(p_1, I_2), I_3; \theta), \\
    \widetilde{p_1} &= \mathcal{F}( \mathcal{F}(\widetilde{p_3}, I_2), I_1 ; \theta).
\end{aligned}
\end{equation}
A longer circle makes the cycle tracking more challenging such that the network has to predict an accurate location of the target object in each single forward and backward tracking. The difference between more forward and backward pairs, such as $p_2$ and $\widetilde{p_2}$, could provide more supervision for the optimization.






\subsection{Siamese Tracking Network}

We follow \cite{li2018high, wang2019fast} to build a Siamese region proposal network, which has a template image patch $\mathbf{z}$ and a search image $\mathbf{x}$ as input. $\mathbf{z}$ is a small patch centered on the target object, and $\mathbf{x}$ is a large patch centered on the last predicted location of the object. These two patches are fed into two fully convolutional subnetworks sharing the same parameters to extract features, after which depth-wise cross-correlation and two branches, box-net, and score-net, are employed to produce a dense response map.
The box-net generates multiple box candidates for each position in the response map. The score-net performs classification and outputs the object and background score for the corresponding box proposals.
Therefore, each element in the response map comprises a set of $k$ box proposals and their corresponding scores. 

Each box proposal encodes 4 normalized coordinates following R-CNN \cite{girshick2014rich}:
\begin{equation}
\begin{aligned}
    t_x &= \frac{x-x_a}{w_a}, t_y=\frac{y-y_a}{h_a}, \\
    t_w &= \log\frac{w}{w_a}, t_h=\log \frac h{h_a},
\end{aligned}
\end{equation}
where $x,y,w, h$ denote the 2-dimensional coordinates of the center, width, and height of the predicted box, while $x_a,y_a,w_a, h_a$ are for the anchor box. 

The score of each box is composed of positive and negative activation $s_{\text{obj}}, s_{\text{back}}$, which are then processed by the softmax function to encode the probability of the box representing an object $p_{\text{obj}}$ and the background $p_{\text{back}}$.

Thus, the output of the region proposal subnetwork is a $4k$ channel box vector and a $2k$ channel score vector, as displayed in Fig.~\ref{fig:network}.

\subsection{Loss Function}

During training, we do not calculate the loss of tracking in the middle of the circle. After a whole circle of the forward and backward tracking, we calculate the loss between the prediction and the initial target. 

The box localization loss for each box is calculated with smooth $L_1$ loss and is formulated as
\begin{equation}
\begin{aligned}
    \mathcal{L}_{\text{box}}=l_1(t_x-t_x^*) + l_1(t_y-t_y^*) + l_1(t_w-t_w^*) + l_1(t_h-t_h^*)
\end{aligned},
\end{equation}
where 
\begin{equation*}
l_1(x)=
\begin{cases}
\frac{1}{2}x^2  \ &|x|<1   \\
|x|-\frac{1}{2}   \ &otherwise
\end{cases} \ ,
\end{equation*}
and $t^*$ is the target box label.

The object score loss for each box is calculated by cross-entropy loss and formulated as
\begin{equation}
\begin{aligned}
    \mathcal{L}_{\text{sco}}=-[y_{\text{o}}\log(p_{\text{obj}})+(1-y_{\text{o}})\log(1-p_{\text{obj}})+ \\
    y_{\text{b}}\log(p_{\text{back}})+(1-y_{\text{b}})\log(1-p_{\text{back}})] 
\end{aligned}\ ,
\end{equation}
where $y_{\text{o}}$ and $y_{\text{b}}$ are the object label and the background label.

Then, the final loss is a weighted sum of these two branches, as
\begin{equation}
    \mathcal{L}=\mathcal{L}_{\text{sco}} + \lambda_1 \mathcal{L}_{\text{box}},
\end{equation}
where the $\lambda_1$ is a weighting factor. With this loss, the cycle tracking framework can be optimized by the self-supervision without the need for expensive annotations.  
\begin{figure}[]
\centering
\includegraphics[width=1.0\columnwidth, trim={0cm 0cm 0cm 0cm}, clip]{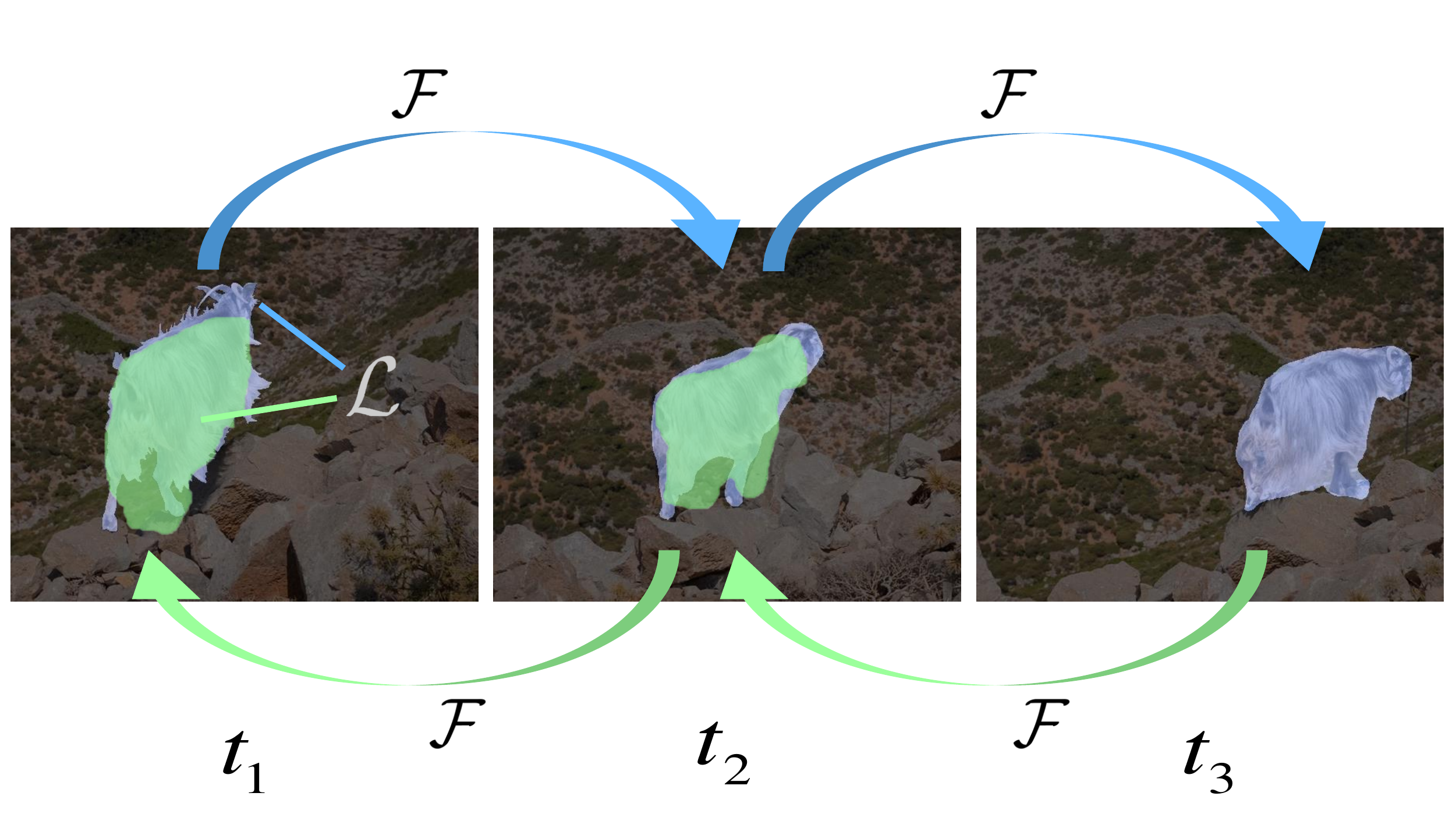}
\caption{Illustration of the self-supervised cycle segmentation propagation framework. Given the blue mask of the target object in the frame, the mask is forward propagated to subsequent frames and then circularly propagated back to the initial frame, generating the predicted mask in green. The forward propagation is indicated by blue arrows while the backward propagation is denoted by green.}
\label{fig:cyclemask}
\end{figure}

\section{SELF-SUPERVISED SEGMENTATION PROPAGATION}

Our cycle-consistent tracking can be extended beyond bounding box tracking. In this section, we first extend the idea for visual object tracking to video segmentation propagation in a self-supervised manner, after which the mask prediction branch is added to assist the segmentation tracking task.

\subsection{Cycle Mask Propagation}

With a similar idea in cycle object tracking, we can address the tracking problem at the mask level, i.e., video object segmentation propagation. 
Given a frame $I_{1}$ at time $t_1$, and the mask of the target object to track, we can first forward propagate the mask to frame $I_{3}$ at time $t_3$, and then backward propagate the mask to frame $I_1$ circularly. Then, between the initial mask and the predicted mask in frame $I_1$, we can obtain a consistency loss as the supervision of this cycle propagation flow, as illustrated in Fig.~\ref{fig:cyclemask}. Therefore, the network can be trained without the need for annotations of every frame. 

\subsection{Siamese Mask Network}

To facilitate the mask propagation in practical applications, following the idea of \cite{wang2019fast}, the input of the mask propagation network is the same as the box tracking network, i.e., a template image patch $\mathbf{z}$ and a search image $\mathbf{x}$ as input. After the depth-wise cross-correlation, now there are three branches for this network: a score-subnetwork, a box-subnetwork, and an additional mask-subnetwork, as shown in Fig.~\ref{fig:network}. With the tensor after cross-correlation as input, for each position of the response map, the mask-net outputs a flattened vector of size $w^m \times h^m$, representing a mask prediction with width $w^m$ and height $h^m$. This mask is resized to the shape of the original search image in inference.

After one propagation circle, the mask loss is calculated for each mask candidate in the response map \cite{wang2019fast} by
\begin{equation}
\mathcal{L}_{\text{mask}}=\sum_n ( \frac{1+y_n}{2w^m h^m} \sum_{i,j} \log (1+e^{-c_n^{ij}m_n^{ij}})), 
\end{equation}
where $m_n^{ij} \in \{\pm1\}$ is the predicted mask label for pixel $(i,j)$ of $n$-th mask candidate, $c_n^{ij}$ denotes the label of the target, and $y_n \in \{\pm1\}$ denotes if the element in the response map is positive. The element is considered positive ($y_n=1$) if one of its $k$ boxes has more than $0.6$ IoU with the target box. 

Then, in the mask propagation network, the final loss consists of three terms:
\begin{equation}
    \mathcal{L}=\mathcal{L}_{\text{sco}} + \lambda_1 \mathcal{L}_{\text{box}} + \lambda_2 \mathcal{L}_{\text{mask} },
\end{equation}
where $\lambda_1$ and $\lambda_2$ are two weighting factors.

\begin{figure*}[]
\centering
\begin{subfigure}{2\columnwidth}
  \centering
  \includegraphics[width=1\columnwidth, trim={0cm 0cm 0cm 0cm}, clip]{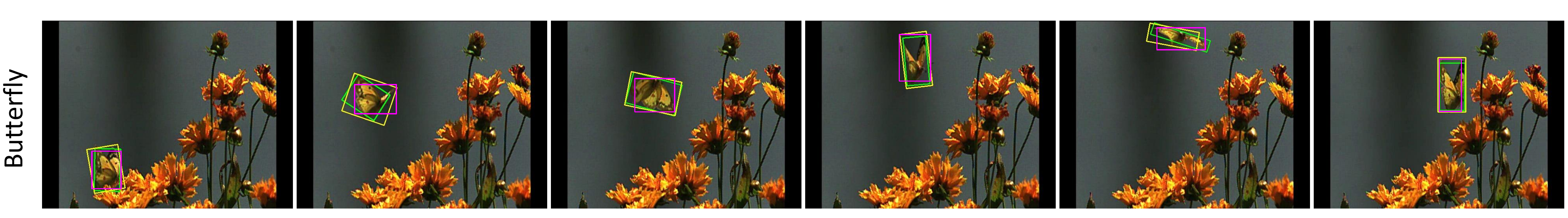}
\end{subfigure}
\begin{subfigure}{2\columnwidth}
  \centering
  \includegraphics[width=1\columnwidth, trim={0cm 0cm 0cm 0cm}, clip]{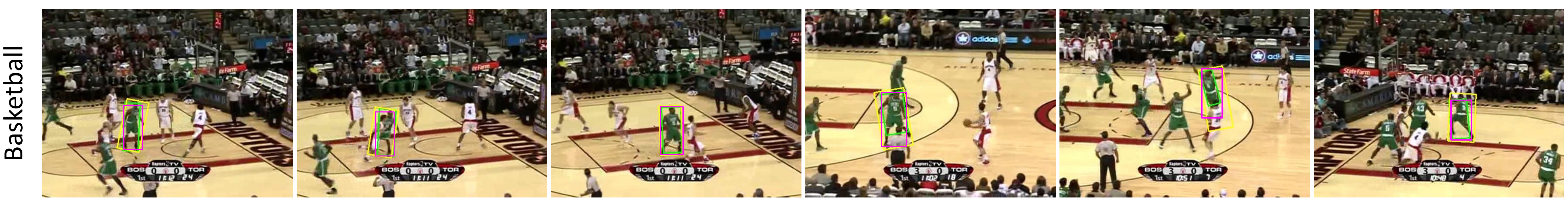}
\end{subfigure}
\begin{subfigure}{2\columnwidth}
  \centering
  \includegraphics[width=1\columnwidth, trim={0cm 0cm 0cm 0cm}, clip]{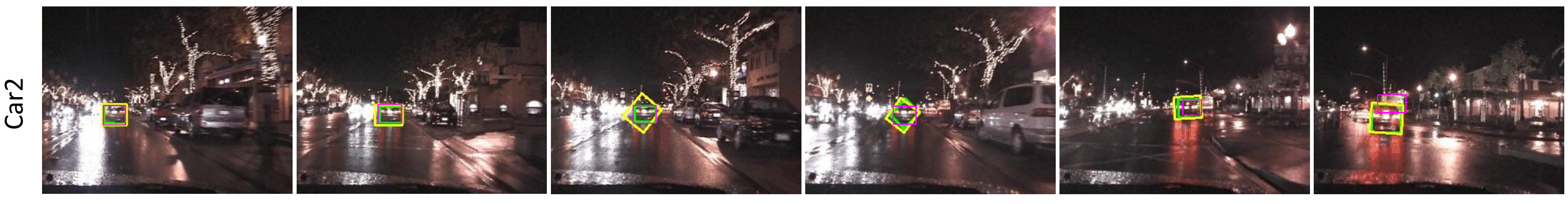}
\end{subfigure}
\begin{subfigure}{2\columnwidth}
  \centering
  \includegraphics[width=1\columnwidth, trim={0cm 0cm 0cm 0cm}, clip]{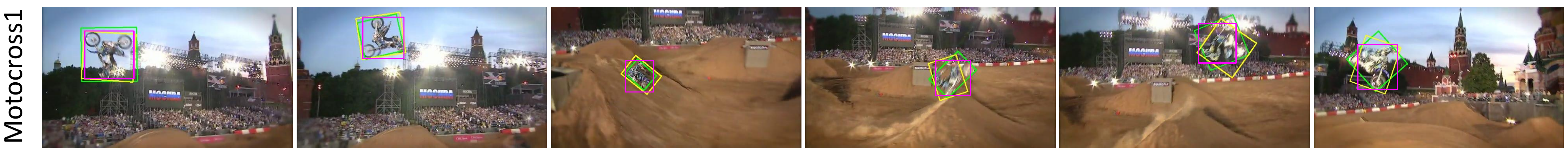}
\end{subfigure}
\caption{Qualitative results of visual object tracking on the VOT dataset. The ground-truth bounding box in green illustrates the target object to track. The predictions of CycleSiam and CycleSiam$^+$ are denoted by pink and yellow boxes, respectively. }
\label{fig:vot}
\end{figure*}

\section{EXPERIMENTS}
\label{sec:exp}

In this section, we first describe in details how we implement the cycle tracking framework and then present the quantitative and qualitative experiments to evaluate our method on both the visual object tracking task and the video object segmentation propagation task.

\subsection{Implementation details}

The self-supervised cycle tracking framework is implemented in Pytorch and trained on one Nvidia GTX 1080 Ti GPU. The batch size is set to $32$, and the weights $\lambda_1$ and $\lambda_2$ are set to $1$ and $30$ without careful searching. The network is optimized end-to-end with SGD, where the learning rate is $0.001$. The input image size for template $\mathbf{x}$ and search $\mathbf{z}$ is $255\times255$ and $127\times127$, respectively. 

In the training, the anchor number $k$ is set to $5$ with ratios $[0.33, 0.5, 1, 2, 3]$ and scale $8$. The object label is defined at where the anchors have $IoU > 0.6$ with the corresponding ground-truth box, and the background label is defined at where the anchors have $IoU < 0.3$ with the ground-truth. The score losses for other anchors are ignored.

We only calculate the loss and optimize the network parameters after one whole cycle of the tracking is finished. In the middle of the cycle tracking, we do not calculate the loss and simply regard it as an inference. For inference, we crop the last predicted box as the template patch $\mathbf{z}$, and crop the image centered on the last prediction as the search patch $\mathbf{x}$. The output box and mask are selected according to the maximum score in the score classification branch. For the mask branch, after a per-pixel sigmoid is applied, we binarize the mask with the threshold of 0.5 to get the final mask output.

\begin{table}[]
\centering
\begin{tabular}{r c c c c}
\toprule
Method & Accuracy & Robustness & EAO & Speed (fps) \\
\midrule
SCT \cite{choi2016visual} & $0.462$ & $0.545$ & $0.188$ & $40$ \\
DSST \cite{danelljan2014accurate} & $0.533$ & $0.704$ & $0.181$ & $25$\\
KCF \cite{henriques2014high} &  $0.489$ & $0.569$ & $0.192$ & $170$  \\
UDT \cite{wang2019unsupervised} & $0.54$ & $0.475$ & $0.226$ & $70$ \\ 
UDT+ \cite{wang2019unsupervised} & $0.53$ & $0.308$ & $0.301$ & $55$ \\ 
CycleSiam  & $\mathbf{0.603}$ & $0.294$ & $0.371$ & $59$ \\ 
CycleSiam$^+$  & $0.601$ & $\mathbf{0.247}$ & $\mathbf{0.398}$ & $44$ \\ 
\bottomrule
\end{tabular}
\caption{Quantitative results of visual object tracking on VOT-2016. Accuracy, robustness, EAO, and speed are reported.}
\label{tab:vot2016}
\end{table}

\begin{table}[]
\centering
\begin{tabular}{r c c c c}
\toprule
Method & Accuracy & Robustness & EAO & Speed (fps) \\
\midrule
DSST \cite{danelljan2014accurate} & $0.395$ & $1.452$ & $0.079$ & $25$\\
KCF \cite{henriques2014high} &  $0.447$ & $0.773$ & $0.135$ & $170$  \\
HMMTxD \cite{vojir2016online} & $0.506$ & $0.815$ & $0.168$ & $-$ \\
CycleSiam  & $\mathbf{0.562}$ & $0.389$ & $0.294$ & $59$ \\ 
CycleSiam$^+$  & $0.549$ & $\mathbf{0.314}$ & $\mathbf{0.317}$ & $44$ \\
\bottomrule
\end{tabular}
\caption{Quantitative results of visual object tracking on VOT-2018.}
\label{tab:vot2018}
\end{table}

\begin{figure*}[]
\centering
\begin{subfigure}{2\columnwidth}
  \centering
  \includegraphics[width=1\columnwidth, trim={0cm 0cm 0cm 0cm}, clip]{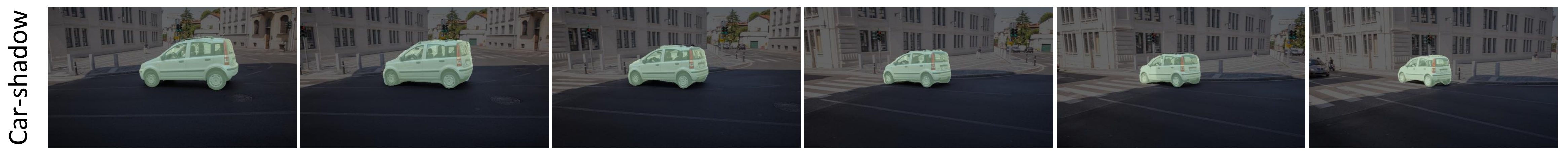}
\end{subfigure}
\begin{subfigure}{2\columnwidth}
  \centering
  \includegraphics[width=1\columnwidth, trim={0cm 0cm 0cm 0cm}, clip]{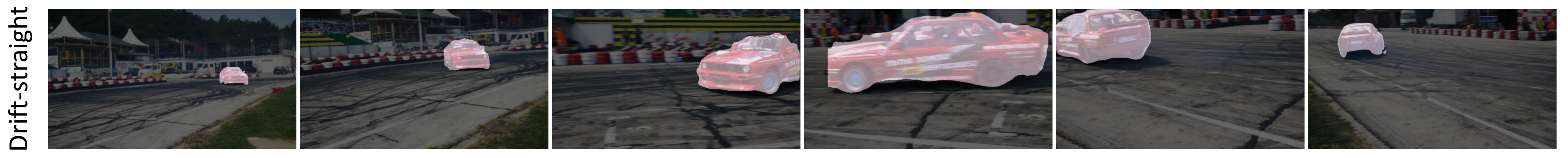}
\end{subfigure}
\begin{subfigure}{2\columnwidth}
  \centering
  \includegraphics[width=1\columnwidth, trim={0cm 0cm 0cm 0cm}, clip]{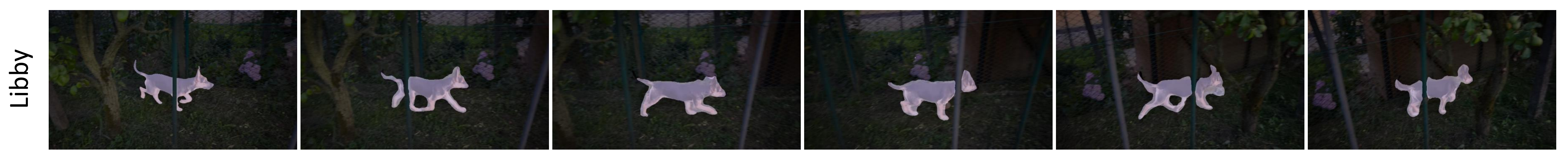}
\end{subfigure}
\begin{subfigure}{2\columnwidth}
  \centering
  \includegraphics[width=1\columnwidth, trim={0cm 0cm 0cm 0cm}, clip]{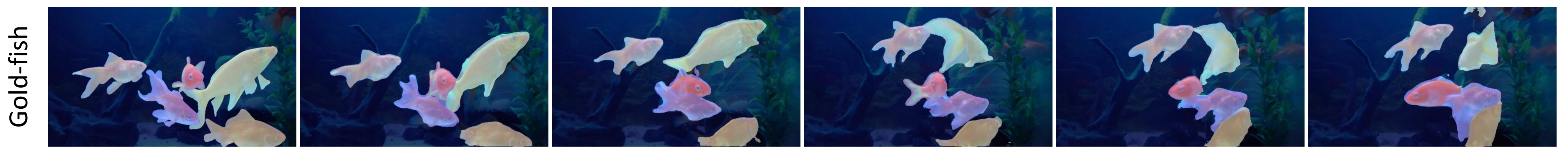}
\end{subfigure}
\caption{Qualitative results of instance mask propagation on the DAVIS-2017 dataset. The first image of every sequence is the input with ground-truth annotation. Different instances are denoted with different colors. Given only the coarse box of the object, our network can predict its instance mask in the subsequent frame in real time.}
\label{fig:davis}
\end{figure*}

\subsection{Self-supervised Visual Object Tracking}

\textbf{Training.} For a fair comparison with existing tracking methods, we train our model on the ILSVRC-2015 dataset. The object box to be tracked is set to be the initial target of the cycle tracking framework, then forward and backward tracking is performed to get the predicted bounding box in the first frame. Between the prediction and initial target object, we calculate the loss and optimize the network. 


\textbf{Evaluation.} After the training, we evaluate our self-supervised object tracking framework on the VOT-2016 and VOT-2018 datasets \cite{kristan2018sixth} without fine-tuning. Both datasets include 60 video sequences, and there are hundreds of frames in each sequence. Some example sequences are shown in Fig.~\ref{fig:vot}. Beginning with the first frame, we run the tracking network once per frame according to the official policy.

The quantitative results are summarized in TABLE~\ref{tab:vot2016} and
TABLE~\ref{tab:vot2018}. We report the running speed and three official metrics, accuracy, robustness, and expected average overlap (EAO).
Since there are few self-supervised deep methods for visual object tracking, we also include some traditional trackers. From the results, we can see that our CycleSiam reaches the EAO of 0.371 in the VOT-2016 dataset and outperforms the previous state-of-the-art self-supervised method UDT by a large margin.

The qualitative results are displayed in Fig.~\ref{fig:vot}. Box predictions of different setup are denoted by different colors. Our method can track the object stably even when the target is mixed with other objects, e.g., the second and fourth images of the Basketball sequence, or when the video is quite blurry, e.g., the last three images of the Motocross sequence.
Sometimes our tracker can give the prediction that is more reasonable than the ground-truth label, e.g., the fourth and fifth image of the Basketball sequence, and the first image of the Motocross sequence. 


\begin{table}[]
\centering
\begin{tabular}{r c c c c}
\toprule
Initialization & Dataset & Accuracy & Robustness & EAO \\
\midrule
\multirow{2}{*}{Random} & VOT-2016 & $0.540$ & $0.735$ & $0.191$  \\ 
& VOT-2018 & $0.377$ & $0.750$ & $0.131$  \\ 
\midrule
\multirow{2}{*}{Object} & VOT-2016 & $0.603$ & $0.294$ & $0.371$  \\ 
& VOT-2018 & $0.562$ & $0.389$ & $0.294$  \\ 
\bottomrule
\end{tabular}
\caption{Ablation study with random target initialization.}
\label{tab:ablation}
\end{table}

\textbf{Ablation study.}
In addition, to test if our framework can work in arbitrary video sequences, we perform an ablation study with random target initialization. The initial target box is no longer given by the tracking target object input, but a random box in the image. In this case, the randomly set target box may contain multiple objects, meaningless background, or parts of an object. This means sometimes the network can be confused by these training data. The performance of our model trained on these messy data is reported in TABLE.~\ref{tab:ablation}. Although the performance is lower, it can still work well in most videos. But the model is confused in some difficult frames where the target object is mixed with the background or other distraction objects. Also, the accuracy is limited, e.g., the predicted box cannot fit tightly to the ground-truth.

\begin{table}[]
\centering
\begin{tabular}{r c c c}
\toprule
Method & $\mathcal{J}$(Mean) & $\mathcal{F}$(Mean) & Speed (fps) \\
\midrule
FCP \cite{perazzi2015fully} & $58.4$ & $49.2$ & $-$ \\ 
BVS \cite{marki2016bilateral} & $60.0$ & $58.8$ & $3$ \\ 
CycleSiam$^+$  & $\mathbf{64.9}$ & $\mathbf{62.0}$ & $31$ \\
\bottomrule
\end{tabular}
\caption{Quantitative results of video object segmentation propagation on DAVIS-2016.  $\mathcal{J}$ is the Jaccard index and $\mathcal{F}$ is the contour F-measure. }
\label{tab:davis2016}
\end{table}

\begin{table}[]
\centering
\begin{tabular}{r c c c}
\toprule
Method & $\mathcal{J}$(Mean) & $\mathcal{F}$(Mean) & Speed (fps) \\
\midrule
SIFT Flow \cite{liu2010sift} &  $33.0$ & $35.0$ & $-$  \\
Transitive-Inv \cite{wang2017transitive} & $32.0$ & $26.8$ & $-$ \\
DeepCluster \cite{caron2018deep} & $37.5$ & $33.2$ & $-$ \\ 
Wang et al. \cite{wang2019learning} & $41.9$ & $39.4$ & $-$ \\
Lai et al. \cite{lai2019self} & $48.4$ & $52.2$ & $-$ \\
CycleSiam$^+$  &  $\mathbf{50.9}$ & $\mathbf{56.8}$ & $31$\\ 
\bottomrule
\end{tabular}
\caption{Quantitative results of video object segmentation propagation on DAVIS-2017.}
\label{tab:davis2017}
\end{table}

\subsection{Self-supervised Video Object Segmentation Propagation}

\textbf{Training.} 
For the video segmentation propagation task, we use the network with three branches.
The propagation network is trained on the YouTube-VOS dataset \cite{xu2018youtube}. The mask of the target object to be propagated in the first frame is set to be the target mask initialization.
Then, the axis-aligned bounding box is extracted from the mask, as the input of our network. Afterward, the forward and backward propagation is performed to get the predicted mask in the first frame. Between the prediction and the original mask, we calculate the loss and optimize the network.

\textbf{Evaluation.} 
After the training, we evaluate our model on the video object segmentation task on the DAVIS-2016 and DAVIS-2017 \cite{pont20172017} validation set without fine-tuning. Given the initial instance mask of the first frame, we extract the axis-aligned bounding box and track the mask in subsequent frames in turn. Similar to most video object segmentation approaches, for multiple instance cases, multiple inferences are performed at the same time. 

The performances of our method, denoted by CycleSiam$^+$, and of other self-supervised methods are presented in TABLE~\ref{tab:davis2017}. The two official metrics, Jaccard index $\mathcal{J}$ for region similarity and contour F-measure $\mathcal{F}$ for contour similarity, are reported. Since there are few self-supervised methods for video segmentation propagation, we also include some visual feature works and use them to find segmentation correspondence, such as SIFT flow \cite{liu2010sift}, Transitive-Inv \cite{wang2017transitive}, and DeepCluster \cite{caron2018deep}, of which the performance is calculated in \cite{wang2019learning}. Additionally, the performance of some traditional methods like FCP \cite{perazzi2015fully} and BVS \cite{marki2016bilateral} are reported.

From the results, we can see that our method outperforms all previous self-supervised algorithms. Please note that the state-of-the-art methods usually use multiple previous frames (7 frames in \cite{wang2019learning}) as the input to predict the mask in the next frame, while we only use the current single frame.
In addition, we report our online speed. One important advantage of our method is that our method can be run in real time. Also, our method only requires a simple bounding box as input. The qualitative results for some videos are presented in Fig.~\ref{fig:davis}. From the visual results, our method can propagate the instance mask stably even when the object size varies significantly, like the Drift-straight sequence. But the mask prediction is not accurate enough in some videos since our method is trained in a self-supervised manner.

Now that we can track the object at the mask level, we can generate a rotated bounding box from the mask. The minimum bounding rectangle for the mask is generated as the box prediction. We evaluate the boxes generated from masks on the VOT-2016 and VOT-2018 datasets and obtain better performance, as reported in TABLE~\ref{tab:vot2016} and
TABLE~\ref{tab:vot2018}. From the visualization in Fig.~\ref{fig:vot}, the rotated box can fit the ground-truth better since the annotation in the VOT dataset is also a rotated box.


\section{CONCLUSION}
\label{sec:conclusion}

In this work, we exploit the end-to-end Siamese network in cycle tracking to perform better self-supervised learning for visual object tracking and video object segmentation propagation. 
By taking advantage of the cycle consistency in a forward and backward tracking circle, self-supervision can be obtained. For the visual object tracking task, the target object is first forward tracked to subsequent frames and then traced back to the first frame. The loss is obtained from the difference between the initial bounding box and the estimated bounding box in the first frame. For the video object segmentation propagation task, in a similar way, the mask is first forward propagated and then circularly propagated back to the first frame, where the consistency loss is calculated to optimize the network.

To leverage the end-to-end learning of deep networks, we introduce the Siamese region proposal network and mask regression network into the tracker, such that a fast and more accurate tracker can be trained end-to-end. In the evaluation experiments on visual object tracking and video object segmentation propagation benchmark datasets, our method outperforms state-of-the-art self-supervised methods in both tasks. In the visual object tracking task, we outperform previous methods by a large margin. In the video segmentation propagation task, we need only a rough bounding box of the objects in the current frame to infer the mask in the next frame, while other methods often use multiple preceding frames. Additionally, our method can be run in real time. These advantages mean that our method could be useful in more practical applications. 

\section{ACKNOWLEDGEMENT}

This work is supported by the Innovation and Technology Fund of the Government of the Hong Kong Special Administrative Region (Project No. ITS/018/17FP, ITS/104/19FP).

\balance



{\small
\bibliographystyle{IEEEtranN}
\bibliography{ref}
}
\end{document}